%% file: main.tex
\documentclass[runningheads]{llncs}

\usepackage{eccv}

\usepackage{eccvabbrv}

\input{preamble}

\usepackage{hyperref}
\usepackage[capitalize]{cleveref}

\usepackage{orcidlink}

\usepackage[width=122mm,left=12mm,paperwidth=146mm,height=193mm,top=12mm,paperheight=217mm]{geometry}
\begin{document}
\input{sec/0_metadata}
\maketitle

\input{sec/0_abstract}
\input{sec/1_introduction}
\input{sec/2_related}
\input{sec/3_method}
\input{sec/4_results}
\input{sec/5_conclusion}

\section*{Acknowledgements}

This work was supported fully by Unity Technologies, without external funding. We would like to thank the reviewers for their valuable feedback and suggestions.

{
    \bibliographystyle{splncs04}
    \bibliography{mark_library, holo_library_manual, manual_library}
}

\end{document}

%% file: preamble.tex
\usepackage[T1]{fontenc}
\usepackage[utf8]{inputenc}
\usepackage[warn]{textcomp}

\usepackage{fontawesome5}
\usepackage{enumitem} %
\usepackage{overpic} %
\usepackage{color}
\usepackage{microtype} %
\usepackage{xspace}
\usepackage{xfrac}
\usepackage{colortbl}
\usepackage{bm}

\usepackage{caption}
\usepackage{subcaption}
\usepackage{wrapfig}
\usepackage{multirow}

\usepackage{graphicx}
\usepackage{amsmath}
\usepackage{amssymb}
\usepackage{booktabs}
\usepackage{adjustbox}
\usepackage{arydshln}
\usepackage[accsupp]{axessibility}  %
\usepackage{pdflscape}

\usepackage{siunitx}
\usepackage{soul,xcolor}
\setstcolor{red}

\usepackage{tikz}
\usepackage{pgfplots}
\pgfplotsset{compat=1.14}

\usetikzlibrary{shapes}
\usetikzlibrary{calc}
\usetikzlibrary{backgrounds}
\usetikzlibrary{positioning}
\usetikzlibrary{arrows}
\usetikzlibrary{arrows.meta}
\usetikzlibrary{decorations.text}    
\usetikzlibrary{fit}
\usetikzlibrary{spy}
\usetikzlibrary{3d}

\usepackage[skins]{tcolorbox}

\definecolor{turquoise}{cmyk}{0.65,0,0.1,0.3}
\definecolor{purple}{rgb}{0.65,0,0.65}
\definecolor{dark_green}{rgb}{0, 0.5, 0}
\definecolor{orange}{rgb}{0.8, 0.6, 0.2}
\definecolor{red}{rgb}{0.8, 0.2, 0.2}
\definecolor{darkred}{rgb}{0.6, 0.1, 0.05}
\definecolor{blueish}{rgb}{0.0, 0.3, .6}
\definecolor{light_gray}{rgb}{0.7, 0.7, .7}
\definecolor{pink}{rgb}{1, 0, 1}
\definecolor{greyblue}{rgb}{0.25, 0.25, 1}

\definecolor{bestcol}{RGB}{254,196,79}

\definecolor{secondbestcol}{RGB}{255,247,188}

\newif\ifreview
\reviewtrue 

\newif\ifdrafting
\draftingtrue %

\ifdrafting
\newcommand{\mb}[1]{{\color{blueish}#1}} %
\newcommand{\MB}[1]{{\color{blueish}{\bf [MB: #1]}}} %
\newcommand{\Mb}[1]{\marginpar{\tiny{\textcolor{blueish}{#1}}}} %
\newcommand{\sv}[1]{{\color{dark_green}#1}}
\newcommand{\SV}[1]{{\color{dark_green}{\bf [SV: #1]}}}
\newcommand{\Sv}[1]{\marginpar{\tiny{\textcolor{dark_green}{#1}}}}
\newcommand{\sd}[1]{{\color{greyblue}#1}}
\newcommand{\SD}[1]{{\color{greyblue}{\bf [SD: #1]}}}
\newcommand{\Sd}[1]{\marginpar{\tiny{\textcolor{greyblue}{#1}}}}
\newcommand{\mpe}[1]{{\color{purple}#1}}
\newcommand{\MP}[1]{{\color{purple}{\bf [SD: #1]}}}
\newcommand{\Mp}[1]{\marginpar{\tiny{\textcolor{purple}{#1}}}}
\else
\newcommand{\mb}[1]{} 
\newcommand{\MB}[1]{} 
\newcommand{\Mb}[1]{} 
\newcommand{\sv}[1]{} 
\newcommand{\SV}[1]{} 
\newcommand{\Sv}[1]{}
\newcommand{\sd}[1]{} 
\newcommand{\SD}[1]{} 
\newcommand{\Sd}[1]{}
\newcommand{\mpe}[1]{} 
\newcommand{\MP}[1]{} 
\newcommand{\Mp}[1]{}
\fi

\newcommand{\vect}[1]{\bm{#1}}

\makeatletter
\DeclareRobustCommand\onedot{\futurelet\@let@token\@onedot}
\def\@onedot{\ifx\@let@token.\else.\null\fi\xspace}

\def\eg{\emph{e.g}\onedot} 
\def\ie{\emph{i.e}\onedot}

\def\etal{\emph{et al}\onedot}

\makeatother

\newcommand{\inlinesection}[1]{\subsubsection{#1}}

\newcommand{\titlecaptionof}[3]{\captionof{#1}{\textbf{#2.}\xspace#3}}

\usepackage{pifont}

\usepackage{blindtext}

\input{colors}

%% file: colors.tex
\definecolor{mpurple}{RGB}{106,27,154}
\definecolor{mpurplelight}{RGB}{206,147,216}

\definecolor{mblue}{RGB}{40,53,147}
\definecolor{mbluelight}{RGB}{159,168,218}

\definecolor{mteal}{RGB}{0,105,92}
\definecolor{mteallight}{RGB}{128,203,196}

\definecolor{morangelight}{RGB}{255,171,145}

\definecolor{mgrayblue}{RGB}{55,71,79}
\definecolor{mgraybluelight}{RGB}{176,190,197}

\definecolor{mamber}{RGB}{255,143,0}
\definecolor{mamberlight}{RGB}{255,224,130}

\definecolor{mdeeporange}{RGB}{216,67,21}
\definecolor{morange}{RGB}{245,124,0}
\definecolor{myellow}{RGB}{253,216,53}

\definecolor{mgreen}{RGB}{85,139,47}
\definecolor{mgreenlight}{RGB}{174,213,129}

\definecolor{mred}{RGB}{198,40,40}
\definecolor{mredlight}{RGB}{239,154,154}

\definecolor{corange1}{RGB}{254,237,222}
\definecolor{corange2}{RGB}{253,190,133}
\definecolor{corange3}{RGB}{253,141,60}
\definecolor{corange4}{RGB}{230,85,13}
\definecolor{corange5}{RGB}{166,54,3}

\definecolor{cblue1}{RGB}{239,243,255}
\definecolor{cblue2}{RGB}{189,215,231}
\definecolor{cblue3}{RGB}{107,174,214}
\definecolor{cblue4}{RGB}{49,130,189}
\definecolor{cblue5}{RGB}{8,81,156}

\definecolor{cgray1}{RGB}{247,247,247}
\definecolor{cgray2}{RGB}{204,204,204}
\definecolor{cgray3}{RGB}{150,150,150}
\definecolor{cgray4}{RGB}{99,99,99}
\definecolor{cgray5}{RGB}{37,37,37}

\definecolor{cred1}{RGB}{254,229,217}
\definecolor{cred2}{RGB}{252,174,145}
\definecolor{cred3}{RGB}{251,106,74}
\definecolor{cred4}{RGB}{222,45,38}
\definecolor{cred5}{RGB}{165,15,21}

\definecolor{cgreen1}{RGB}{237,248,233}
\definecolor{cgreen2}{RGB}{186,228,179}
\definecolor{cgreen3}{RGB}{116,196,118}
\definecolor{cgreen4}{RGB}{49,163,84}
\definecolor{cgreen5}{RGB}{0,109,44}

\definecolor{cdiv11}{RGB}{166,97,26}
\definecolor{cdiv12}{RGB}{223,194,125}
\definecolor{cdiv13}{RGB}{245,245,245}
\definecolor{cdiv14}{RGB}{128,205,193}
\definecolor{cdiv15}{RGB}{1,133,113}

\definecolor{cdiv21}{RGB}{208,28,139}
\definecolor{cdiv22}{RGB}{241,182,218}
\definecolor{cdiv23}{RGB}{247,247,247}
\definecolor{cdiv24}{RGB}{184,225,134}
\definecolor{cdiv25}{RGB}{77,172,38}

\definecolor{cdiv31}{RGB}{230,97,1}
\definecolor{cdiv32}{RGB}{253,184,99}
\definecolor{cdiv33}{RGB}{247,247,247}
\definecolor{cdiv34}{RGB}{178,171,210}
\definecolor{cdiv35}{RGB}{94,60,153}

\definecolor{bestcol}{RGB}{255,247,188}
\newcommand{\bestcolor}[1]{\cellcolor{bestcol}\textbf{#1}}

%% file: sec/0_metadata.tex
\pagestyle{headings}
\mainmatter{}
\def\ECCVSubNumber{2035}

\title{Collaborative Control for Geometry-Conditioned PBR Image Generation}

\titlerunning{Collaborative Control for Geometry-Conditioned PBR Image Generation}
\authorrunning{S.~Vainer \etal}
\author{Shimon Vainer\inst{1, \star} \and
Mark Boss\inst{2, \dag}\orcidlink{0000-0001-7515-4934} \and Mathias Parger\inst{1} \and Konstantin Kutsy\inst{1} \and Dante De Nigris\inst{1} \and Ciara Rowles\inst{1} \and Nicolas Perony\inst{1} \and Simon Donn\'{e}\inst{1, \star}\orcidlink{0000-0002-7461-8589}}
\institute{%
Unity Technologies\inst{1} \qquad Stability AI, work done while at Unity Technologies\inst{2}\\
$\star$ Equal Contributions \quad $\dag$ Core Technical Contributions \\
Corresponding author: \href{mailto:shimon.vainer@unity3d.com}{shimon.vainer@unity3d.com}
}

%% file: sec/0_abstract.tex
\begin{figure}
    \centering
    \input{fig/teaser_tikz}
\end{figure}
\begin{abstract}
Graphics pipelines require physically-based rendering (PBR) materials, yet current 3D content generation approaches are built on RGB models.
We propose to model the PBR image distribution directly, avoiding photometric inaccuracies in RGB generation and the inherent ambiguity in extracting PBR from RGB.
As existing paradigms for cross-modal fine-tuning are not suited for PBR generation due to both a lack of data and the high dimensionality of the output modalities, we propose to train a new PBR model that is tightly linked to a frozen RGB model using a novel cross-network communication paradigm.
As the base RGB model is fully frozen, the proposed method retains its general performance and remains compatible with \eg IPAdapters for that base model.
\keywords{Image Generation, Material Properties, Multi-Modal Generation, Physically-Based Rendering}
\end{abstract}

%% file: fig/teaser_tikz.tex
\resizebox{\textwidth}{!}{
\begin{tikzpicture}[
    font={\fontsize{8pt}{10}\selectfont},
    textlayer/.style={draw=#1, fill=#1!20, line width=1.5pt, inner sep=0.1cm, rounded corners=1pt},
]

\node[align=center] (snake_normals)  at (0,0) {\includegraphics[width=1.9cm]{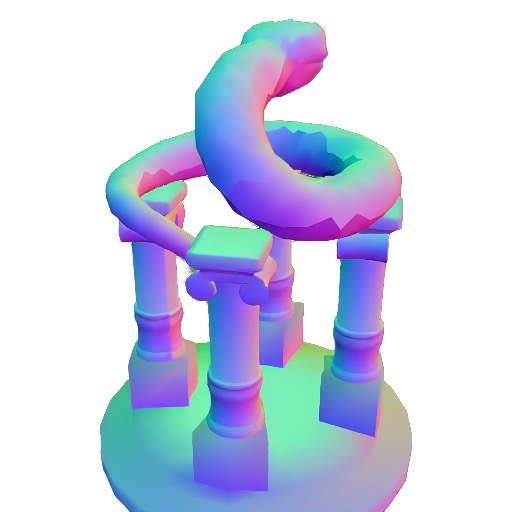}};
\node[align=center] (shield_normals) at (-5.25,0) {\includegraphics[width=1.9cm]{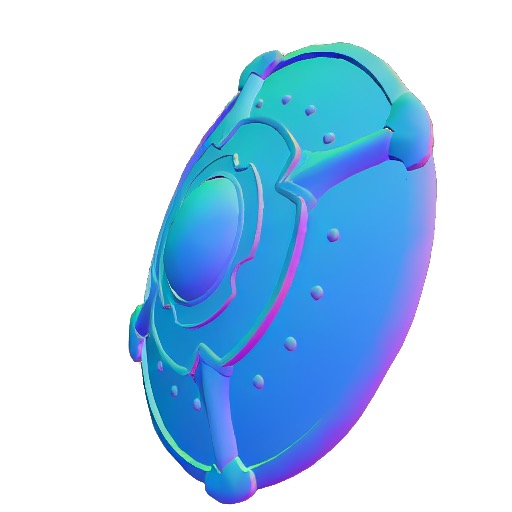}};
\node[align=center] (sphere_normals) at (5.25,0) {\includegraphics[width=1.9cm]{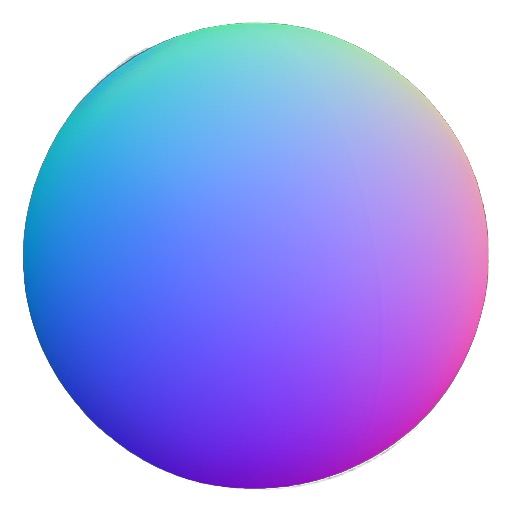}};

\foreach \offset / \channel in {-2/albedo,-4/roughness_metallic,-6/tangent_normals} {
    \node[align=center] (shield_\channel_1) at (-7.25, \offset) {\includegraphics[width=1.9cm]{img/teaser/60_seed_42_\channel.png.jpg}};
    \node[align=center] (shield_\channel_2) at (-5.25, \offset) {\includegraphics[width=1.9cm]{img/teaser/61_seed_1337_\channel.png.jpg}};
    \node[align=center] (shield_\channel_3) at (-3.25, \offset) {\includegraphics[width=1.9cm]{img/teaser/62_seed_42_\channel.png.jpg}};
    
    \node[align=center] (snake_\channel_1) at (-1, \offset) {\includegraphics[width=1.9cm]{img/teaser/32_seed_42_\channel.png.jpg}};
    \node[align=center] (snake_\channel_2) at (1, \offset) {\includegraphics[width=1.9cm]{img/teaser/33_seed_1337_\channel.png.jpg}};
    
    \node[align=center] (sphere_\channel_1) at (3.25, \offset) {\includegraphics[width=1.9cm]{img/teaser/130_seed_42_\channel.png.jpg}};
    \node[align=center] (sphere_\channel_2) at (5.25, \offset) {\includegraphics[width=1.9cm]{img/teaser/132_seed_42_\channel.png.jpg}};
    \node[align=center] (sphere_\channel_3) at (7.25, \offset) {\includegraphics[width=1.9cm]{img/teaser/134_seed_1337_\channel.png.jpg}};
};

\node[align=center, font=\scriptsize] (albedo_text) at (-8.25, 0) {\rotatebox{90}{\parbox{1.5cm}{Input\\Geometry}}};
\node[align=center, font=\scriptsize] (albedo_text) at (-8.25, -2) {\rotatebox{90}{Albedo}};
\node[align=center, font=\scriptsize] (RM_text) at (-8.25, -4) {\rotatebox{90}{\parbox{1.5cm}{Roughness/\\Metallic}}};
\node[align=center, font=\scriptsize] (bump_text) at (-8.25, -6) {\rotatebox{90}{Bump map}};

\end{tikzpicture}
}
\titlecaptionof{figure}{Generated PBR materials}{%
By tightly linking the PBR diffusion model with a frozen RGB model, we produce high-quality PBR images conditioned on geometry and prompts.
Visit the project page at \href{https://unity-research.github.io/holo-gen}{https://unity-research.github.io/holo-gen}.
}\label{fig:teaser}%

%% file: sec/1_introduction.tex
\section{Introduction}\label{sec:introduction}
The recent meteoric rise of diffusion models has made at-scale generation of high-quality RGB image content more accessible than ever and Text-to-Texture and Text-to-3D approaches successfully lift this to 3D~\cite{li2024advances}.
But to maximize the usefulness of the generated textures in downstream 3D workflows, generated content must be compatible with physically-based rendering (PBR) pipelines for proper shading and relighting.
Current approaches rely on generated RGB images and subsequent PBR extraction through inverse rendering, suffering from the physically \textbf{in}accurate lighting in the generated RGB images as well as from significant ambiguities in the inverse rendering.
We propose a solution for geometry-conditioned generation of PBR images by modeling the joint distribution directly, avoiding the issues around photometric consistency and inverse rendering.

To model the distribution of non-RGB modalities, existing approaches typically fine-tune the weights of a base RGB model~\cite{ke2023repurposing,stan2023ldm3d,lee2023exploiting, du2023intrinsic,long2023wonder3d,liu2023unidream}.
Applied to PBR images, this means either directly predicting the entire PBR image stack or sequentially predicting them conditioned on one another.
Neither is sufficient for our use-case: jointly predicting the entire PBR image stack is problematic as the higher-dimensional modality does not compress well into the established latent spaces (as we show in \cref{sec:results}), and sequentially predicting the elements of the PBR image stack is significantly more expensive and risks compounding errors in the sequential generation.
Furthermore, while state-of-the-art RGB diffusion models are trained on billions of images~\cite{schuhmann2022laion}, there is unfortunately no dataset of such size at our disposal for PBR content generation.
Instead, the largest available dataset of PBR content is Objaverse~\cite{deitke2023objaverse}, containing around \num[group-separator={,}]{800000} objects with associated PBR textures, limited to ``everyday'' appearances of the objects.
In light of the restricted training data available, fine-tuning the base model results in catastrophic forgetting, forfeiting generalizability, as we illustrate in \cref{sec:results}.

Instead, we keep a pre-trained RGB image model frozen and train a parallel model to generate PBR images, as shown in \cref{fig:overview}.
We tightly link the PBR model to the frozen RGB model using our proposed cross-network control paradigm, in order to leverage its expressivity and rich internal state.
As a result we are able to generate qualitative and diverse PBR content, even for unlikely appearances of objects (far out-of-distribution for the Objaverse dataset).
Crucially, the frozen RGB model safeguards against catastrophic forgetting \emph{and} remains compatible with techniques such as IPAdapter~\cite{ye2023ip-adapter}.
In summary, we:
\begin{enumerate}[nosep]
    \item Propose the novel \emph{Collaborative Control} paradigm to tightly link the PBR generator to a fully frozen pre-trained RGB model, modeling the joint distribution of RGB and PBR images directly (see \cref{sec:cross-network communicatoin}),
    \item Illustrate that the proposed control mechanism is data-efficient, and generates high-quality images even from a very restricted training set,
    \item Demonstrate the compatibility with IPAdapter~\cite{ye2023ip-adapter} specifically, and
    \item Ablate our design choices to show the improvement over existing paradigms in literature and the issues with existing paradigms.
\end{enumerate}

\begin{figure}[ht]
    \input{img/tikz/overview_tikz}
\end{figure}

%% file: img/tikz/overview_tikz.tex
\begin{minipage}[c]{0.5\linewidth}
    \include{img/tikz/overview_parts/collaborative_control_main}
\end{minipage}
\hfill
\begin{minipage}[c]{0.48\linewidth}
    \include{img/tikz/overview_parts/collaborative_control_comm}
\end{minipage}
\titlecaptionof{figure}{Collaborative Control}{
Two parallel models collaborate to generate pixel-aligned outputs of different modalities.
We freeze the left \textcolor{blueish}{pre-trained RGB model} and train the right \textcolor{orange}{PBR model} with its \textcolor{orange}{cross-network communication layers}.
The cross-communication concatenates the states of both models, processes them with a small MLP, and residually distributes the result back to the respective models.
As discussed in \cref{sec:results}, prompt cross-attention in the PBR model is counter-productive.
}\label{fig:overview}

%% file: sec/2_related.tex
\section{Related Work}\label{sec:related}
\inlinesection{Generating natural images from text prompts.}\label{ssec:related_work_rgb_diffusion}\label{ssec:related_work_multi_modal}
Natural image generation has a long history: from GANs~\cite{goodfellow2020generative, karras2017progressive, karras2019style, karras2020analyzing, karras2021alias} and VAEs~\cite{kingma2013auto}, to autoregressive models~\cite{van2016conditional, van2016pixel}.
More recently, the introduction of diffusion models~\cite{sohl2015deep, ho2020denoising, song2020denoising} was a breakthrough in the generative field --- far more stable than typical GAN training, albeit slow and computationally expensive, and easier to control and condition.
Unfortunately, these approaches require billions of images to train from scratch~\cite{schuhmann2022laion}, and
for PBR image generation the largest commonly available dataset is Objaverse~\cite{deitke2023objaverse}.
With \num[group-separator={,}]{800000}+ objects it is still several orders of magnitude smaller than LAION-5B~\cite{schuhmann2022laion} and proves insufficient to train generative models that can generalize to unlikely semantics (as illustrated in~\cref{sec:results}).
While pre-trained RGB models encode rich prior knowledge around structure, semantics, and materials~\cite{sharma2023alchemist, subias2023wild, du2023intrinsic}, Sarkar \etal warn that the models are often still geometrically inaccurate~\cite{sarkar2023shadows}: we have found that this extends to material properties, as diffusion models prefer idealized and artistic appearances over photometric accuracy.

\inlinesection{Generating non-RGB modalities}
Existing works fine-tune pre-trained RGB models to predict \eg Depth~\cite{ke2023repurposing,stan2023ldm3d}, semantics~\cite{lee2023exploiting} or intrinsic properties~\cite{du2023intrinsic,long2023wonder3d,liu2023unidream}, either directly or through LoRA's~\cite{hu2021lora}.
Sadly, this is not plausible for PBR image generation: compressing PBR images into the existing low-dimensional latent space overloads it, and the alternative of sequentially predicting channel triplets is too costly and slow.
Wonder3D~\cite{long2023wonder3d} and UniDream~\cite{liu2023unidream} perform joint RGB and normal diffusion using a cross-domain self-attention aligning the two parallel branches, yet this scales poorly an increasing number of output modalities.
Our proposed approach instead uses a frozen RGB model, negating the risk of catastrophic forgetting, and trains a parallel branch for all additional modalities jointly (in the latent space of a PBR VAE), reducing cost.

\inlinesection{Image-based conditioning}
Existing pixel-accurate control techniques come in two flavors: re-training of the base model with modified input spaces~\cite{duan2023diffusiondepth, ke2023repurposing}, and training of a parallel model that affects the base model's state~\cite{zhang2023adding,zavadski2023controlnetxs,hu2023animateanyone}.
We find in \cref{sec:results} that the former risks losing the base model's expressiveness and quality.
In ControlNet~\cite{zhang2023adding} and ControlNet-XS~\cite{zavadski2023controlnetxs}, the controlling model only influences the base RGB model's output while in AnimateAnyone~\cite{hu2023animateanyone} the parallel model is only tasked with generating its own output.
Instead, we leave the RGB base model's weights fully frozen and residually edit its internal states from a parallel model that is itself tasked with generating PBR images: our PBR model both \emph{controls} the base model (to guide it towards the domain of rendered images), and \emph{generates} its own PBR output (based on the RGB model's internal state); therefore, our proposed approach requires full bidirectional connections between both branches as shown in \cref{fig:control_diagram}.
To condition on input geometry we concatenate it to the input of the PBR branch, as Ke \etal~\cite{ke2023repurposing}.

\inlinesection{Text-to-3D} describes the task of generating 3D objects from text prompts, often with the aim to support downstream graphics pipelines such as game engines.
Earlier methods leverage Score Distillation Sampling~\cite{poole2022dreamfusion} (SDS) to iteratively optimize a 3D representation by backpropagating the diffusion model's noise predictions~\cite{wang2023steindreamer, ma2023geodream, wu2024consistent3d, tang2023stable, zhou2023dreampropeller, zhuang2023dreameditor, huang2023dreamtime, wang2023prolificdreamer, shi2023mvdream, ma2023x, guo2023stabledreamer, liang2023luciddreamer, wang2023luciddreaming, tang2023dreamgaussian, liu2023sherpa3d} through the RGB model, or building on viewpoint-aware image models~\cite{yu2023boosting3d, raj2023dreambooth3d, liu2023zero, shi2023zero123++, huang2023dreamcontrol, long2023wonder3d, liu2023pi3d} for direct fusion.
Such RGB methods ignore that object appearance varies with viewing angle, often resulting in artifacts around highlights, and their RGB output is not useful in graphics pipelines.
More recent work generates PBR properties so using inverse rendering with a differentiable renderer~\cite{wu2023hyperdreamer, xu2023matlaber, liu2023unidream, chen2023fantasia3d, yeh2024texturedreamer}: a major concern is lighting  being baked into the material channels (\eg HyperDreamer~\cite{wu2023hyperdreamer} uses an ad-hoc regularization to reduce these artifacts).
\textbf{Text-to-Texture} methods restrict the Text-to-3D problem to objects with known structure by conditioning the diffusion model on the object geometry~\cite{le2023euclidreamer, youwang2023paint, zeng2023paint3d, knodt2023consistent, cao2023texfusion, chen2023text2tex, zhang2023repaint123}, but face similar issues by operating in the RGB domain.
Paint3D~\cite{zeng2023paint3d} also discusses the lighting artifacts typical with inverse rendering and introduces a custom post-processing diffusion model to alleviate these.
By directly generating PBR content, our proposed technique promises to resolve issues related to inverse rendering in the latter methods, all the while retaining the simplicity of the former methods.

\inlinesection{Evaluation metrics for generative methods} compare the output distributions with known ground-truth distributions, typically with the Inception Score~\cite{salimans2016improved} (IS) or the Fréchet Inception Distance~\cite{heusel2017gans} (FID).
CMMD~\cite{jayasumana2023rethinking} argues that neither is well suited to modern generative models, and compare the distributions of CLIP embeddings rather than Inceptionv3~\cite{szegedy2016rethinking} internal states.

Aside from comparing modelled distributions with the ground truth, we also wish to evaluate the alignment of the generated images to their text prompts.
CLIPScore~\cite{hessel2021clipscore} compares the image's CLIP embedding with that of the prompt: whether all the relevant elements are represented and whether any extraneous elements were introduced.
We also report the OneAlign \emph{aesthetics} and \emph{quality} metrics of the generated images~\cite{wu2023qalign}, which have been shown to align well with human perception, to provide a more quantitative indication of quality.

%% file: sec/3_method.tex
\section{Preliminaries}\label{sec:problem_statement}
\inlinesection{PBR materials} are a compact representation of the bidirectional reflectance distribution function (BRDF), which describes how light is reflected from the surface of an object.
We use the popular Cook-Torrance analytical BRDF model~\cite{Cook1982}, using specifically the Disney BRDF Basecolor-Metallic parametrization~\cite{Burley2012} as it inherently promotes physical correctness.
In this parametrization, the BRDF comprises \textit{Albedo} ($\vect{b}_a \in \mathbb{R}^{3}$), \textit{Metallic} ($\vect{b}_m \in \mathbb{R}$), and \textit{Roughness} ($\vect{b}_r \in \mathbb{R}$) components.
To increase realism during rendering beyond the resolution of the underlying geometry (often a mesh), graphics pipelines add small details such as wood grain or grout between tiles by encoding them as offsets to the surface normals in an additional \textit{bump map} ($\vect{b}_n \in \mathbb{R}^{3}$).
As this bump map is typically defined in a tangent space based on an arbitrary UV-unwrapping, it entangles the surface property with this arbitrary UV mapping.
Instead, we propose to predict the bump map defined in a tangent space based solely on the object geometry,  disentangling the texture from the UV mapping as shown in \cref{fig:markspace_normal}.
To construct this geometry tangent space for a point $\vect{p} = \left[p_x, p_y, p_z\right]^T$ with geometry normal $\vect{n}$, we construct the local tangent vector as $\vect{t} = \vect{n} \times (\left[-p_y, p_x, 0\right]^T\times \vect{n})$, corresponding to Blender's \emph{Radial Z} geometry tangent.
The geometry tangent space is then constructed as $\left(\vect{t} / \Vert\vect{t}\Vert, \vect{n} \times \vect{t} / \Vert\vect{t}\Vert, \vect{n}\right)^T$.

\begin{figure}[t]
    \centering
    \begin{minipage}[b]{0.55\textwidth}
        \input{fig/markspace_example}
    \end{minipage}\hfill
    \begin{minipage}[b]{0.4\textwidth}
        \input{fig/static_illumination}
    \end{minipage}
\end{figure}

\inlinesection{Diffusion models}~\cite{sohl2015deep,ho2020denoising} iteratively invert a forward degradation process to generate high-quality images from pure noise (typically white Gaussian noise).
Formally, the forward process iteratively degrades images from the data distribution $\vect{z}_0 \sim p(\vect{z})$ to standard-normal samples $\vect{z}_T \sim \mathcal{N}(0, I)$ over the course of T degradation steps as $\vect{z}_t \sim \mathcal{N}(\alpha_t \vect{z}_{t-1}, (1 - \alpha_t) I)$,
where $\alpha_{t}$ denotes the noise schedule for timestep $t$.
Practically, the forward process can be condensed into the direct distribution $\vect{z}_t \sim \mathcal{N}(\sqrt{\bar{\alpha}_t} \vect{z}_{0}, (1 - \bar{\alpha}_t) I)$ with the appropriate choice of  $\bar{\alpha}_t$.
The diffusion model $\mathcal{D}$ is trained to sample the stochastic reverse process $\mathcal{D}_{t}(\vect{z}_{t}) \sim p(\vect{z}_{t-1} \vert \vect{z}_{t})$ to iteratively generate $\vect{z}_{0}$ from $\vect{z}_{T}$.

\section{Method}\label{sec:method}
We wish to train a PBR diffusion model $\mathcal{D}_{pbr}$ that models the reverse denoising process for PBR images as represented in the latent space of a VAE~\cite{rombach2022high}, representing the data distribution $p(\vect{z}_{pbr})$.
We find that we lack the data required to train this model directly, and instead propose to model $p(\vect{z}^{\prime}_{rgb}:=f_{rgb}(\vect{z}_{pbr}), \vect{z}_{pbr})$ based on an RGB diffusion model $\mathcal{D}_{rgb}$ for the RGB data distribution $p(\vect{z}_{rgb})$; $f_{rgb}$ is a rendering function that projects the PBR images onto the RGB domain.
To motivate this, we split the joint reverse process into two separate processes using Bayes' rule:
\begin{equation}
\begin{aligned}
    p(&\vect{z}^{\prime}_{rgb, t-1}, \vect{z}_{pbr, t-1} \vert \vect{z}^{\prime}_{rgb, t}, \vect{z}_{pbr, t}) \\
    &\sim \textcolor{blue}{p(\vect{z}^{\prime}_{rgb, t-1} \vert \vect{z}^{\prime}_{rgb, t}, \vect{z}_{pbr, t})} \textcolor{orange}{p(\vect{z}_{pbr, t-1} \vert \vect{z}^{\prime}_{rgb, t-1}, \vect{z}^{\prime}_{rgb, t}, \vect{z}_{pbr, t})}
\end{aligned}
\label{eq:joint_to_parallel_diffusion}
\end{equation}
The \textcolor{blue}{RGB model} is implemented based on $\mathcal{D}_{rgb}(\vect{z}_{rgb, t-1}) \sim p(\vect{z}_{rgb, t-1} \vert \vect{z}_{rgb, t})$: we align the current RGB sample with the PBR sample and restrict it to $\mathrm{Im}(f)$ (the domain of rendered images with the fixed environment map) so that its internal states are more easily interpreted by the PBR branch\footnote{Our intuition as to why a fixed environment map is beneficial is that it makes the RGB model's internal states more consistent to interpret, and makes the control problem of projecting to $\mathrm{Im}(f)$ simpler. Early in training, generated sample quality can be boosted significantly by applying the foreground mask to the RGB estimate for the first few timesteps; a rough projection to bring the estimate much closer to $\mathrm{Im}(f)$. After longer training, this is no longer necessary, as the PBR branch is capable enough to restrict the RGB branch to $\mathrm{Im}(f)$.}.
To simplify this alignment problem, the rendering function $f_{rgb}$ uses fixed camera settings and a fixed environment map as shown in \cref{fig:camera_colocated_env_illumination}.
The \textcolor{orange}{PBR model} now no longer models $p(\vect{z}_{pbr, t-1} \vert \vect{z}_{pbr, t})$: it additionally has access to the RGB context $(\vect{z}^{\prime}_{rgb, t-1}, \vect{z}^{\prime}_{rgb, t})$ which simplifies the problem.
The RGB and PBR models are in practice much more intertwined than \cref{eq:joint_to_parallel_diffusion} implies: this derivation serves mostly as an intuitive indication for why the joint problem is more tractable.
Note that $\vect{z}^{\prime}_{rgb, t}$ is a degraded version of $\vect{z}^{\prime}_{rgb, 0}$, and \emph{not} a rendered version of $\vect{z}_{pbr, t}$: the PBR model does not learn to do inverse rendering in degraded image space but rather learns to denoise PBR images given additional RGB context.

\subsection{Collaborative Control}\label{sec:cross-network communicatoin}
In summary, our proposed approach comprises two models working in tandem: a pre-trained RGB image model and a new PBR model (see \cref{fig:control_diagram} for a high-level overview of our proposed control scheme). The previous section identifies two tasks for this cross-network communication: aligning the RGB model's output with both the PBR model's output and the map of the rendering function $\mathrm{Im}(f)$, and communicating knowledge in the RGB model to the PBR model.
ControlNet~\cite{zhang2023adding} and ControlNet-XS~\cite{zavadski2023controlnetxs} discuss solutions to the former control problem --- the authors conclude that communication from the base model's encoder to the controlling model's encoder, and from the controlling model's decoder to the base model's decoder, is sufficient.
AnimateAnyone~\cite{hu2023animateanyone} addresses the latter problem and concludes that, there, uni-directional communication from the left model to the right model is sufficient.
We have found that full bidirectional communication is crucial for our approach: the PBR branch needs to extract relevant information from the RGB model's hidden state, while simultaneously guiding the RGB output towards render-like images (\ie{} with a black background and compatible lighting) to ensure those hidden states are consistent with its own expectations.
We dub this \emph{Collaborative Control}.

\begin{figure}[t]
    \input{img/tikz/control_diagram_tikz.tex}
\end{figure}

We implement the cross-network communication as a connecting layer between the two models after every self-attention module; its inputs are the concatenation of the model states and its outputs are residually distributed to both models again.
During training, we only optimize the weights of the PBR model and the cross-network communication links against both models' outputs, while the RGB model remains fully frozen.
By adopting this approach, we safeguard the base RGB model's weights, and do not risk catastrophic forgetting for that base model.
As we discuss in \cref{sec:results}, we have found that a single per-pixel linear layer is sufficient, although we also evaluate the other control schemes from \cref{fig:control_diagram} as well as an attention-based communication layer.
Notably, we have also found that disabling the text cross-attention in the PBR model is crucial to out-of-distribution performance; we attribute this to overfitting on the restricted dataset, as this problem worsens with reduced training data.
Only allowing prompt attention through the frozen RGB model prevents such overfitting.

\subsection{Implementation}\label{sec:joint_generation}
\inlinesection{Compressing PBR images into latent space}
RGB diffusion models benefit immensely from a dedicated VAE to encode the images into a lower-dimensional latent space~\cite{rombach2022high}.
Existing solutions that generate an alternate modality typically encode that modality with the RGB VAE, but PBR images cannot be compressed into the same latent space due to the higher dimensionality.
Instead, we could select channel triplets $\vect{b}_a$, $\left[\vect{b}_m, \vect{b}_r, \vect{0}\right]$, and $\vect{b}_n$ and process those with the RGB VAE, but we instead choose to train a dedicated PBR VAE --- our ablation studies indicate that the distribution mismatch between the PBR channels and the RGB space is too large, and performance would otherwise suffer.
We adopt the VAE architecture and training code from StableDiffusion v1.5~\cite{rombach2022high}, although following Vecchio et al.~\cite{vecchio2023controlmat} we set the latent space channel count to $14$ for the optimal balance between quality and compression when processing PBR images.

\inlinesection{Conditioning on existing geometry}
We concatenate the screen-space geometry normals to the PBR model's inputs to condition the joint output.
Referring to \cref{fig:control_diagram}, Collaborative Control encapsulates the ControlNet scheme that would typically be used for this conditioning~\cite{zavadski2023controlnetxs}: as we jointly train from scratch, this does not introduce additional cost.

\inlinesection{Generating training data}
Our dataset for training both the PBR VAE and the Collaborative Control scheme is based on Objaverse~\cite{deitke2023objaverse}: a dataset containing \num[group-separator={,}]{800000}+ 3D models with annotations for what the models represent (describing both shape and texture).
After sanitizing and filtering the dataset we retain roughly \num[group-separator={,}]{300000} objects.
Each of the objects is rendered with Blender from 16 viewpoints encircling the object using a fixed pinhole camera model and a fixed (camera colocated) environment map\footnote{\url{https://polyhaven.com/a/studio_small_08}} as in \cref{fig:camera_colocated_env_illumination}.
For the evaluations in \cref{sec:results}, we leave out 2\% randomly selected elements.

\inlinesection{Training Collaborative Control}
For most of the experiments in \cref{sec:results}, ZeroDiffusion~\cite{lin2024zerosnr,drhead2024ZeroDiffusion} is the base RGB model, a zero-terminal-SNR version fine-tuned from StableDiffusion v1.5~\cite{rombach2022high}.
As Collaborative Control is agnostic to the base model, we also illustrate StableDiffusion v1.5 and v2.1 as base models in \cref{sec:results}.
We optimize the PBR model's weights as well as the cross-network communication layers to minimize the training loss for the RGB and PBR denoising jointly, while keeping the RGB model fully frozen.
Unless otherwise stated, we directly train on $512 \times 512$ resolution for a total of \num[group-separator={,}]{200000} update steps with a batch size of $12$ and a learning rate of $3\times 10^{-5}$ (on one $80$ GB VRAM A100, taking roughly two days).
We evaluate the effect of a larger training budget by training on $8$ A100's for the same number of steps, increasing the batch size by a factor of $8$ without affecting training time --- for environmental and cost purposes, the training budget is kept low for the main ablation study.

%% file: fig/markspace_example.tex
\begin{center}%
\resizebox{\linewidth}{!}{
\renewcommand{\arraystretch}{0.6}%
\setlength{\tabcolsep}{0pt}%
\begin{tabular}{%
    @{}>{\centering\arraybackslash}m{4cm}%
    >{\centering\arraybackslash}m{4cm}%
    >{\centering\arraybackslash}m{4cm}%
    @{}
}
World space & UV tangent space & Geometry Tangent Space \\

\begin{tikzpicture}[spy using outlines={circle, magnification=3, connect spies}]
    \node { \includegraphics[width=4cm]{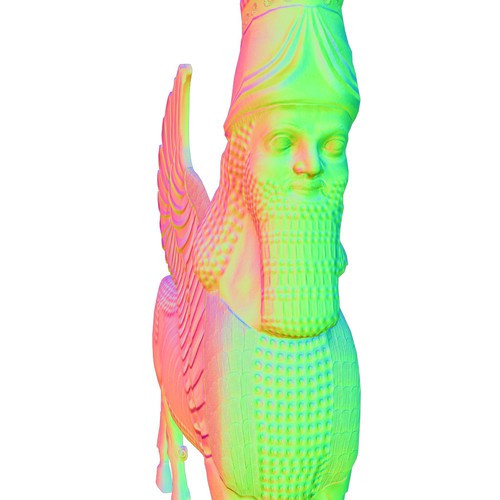} };
    \spy[size=1.25cm] on (-0.5, -1.0) in node at (1.35, -0.5);
    \spy[size=1.25cm] on (-0.6, 0.5) in node at (1.35, 1.5);
\end{tikzpicture} 
&
\begin{tikzpicture}[spy using outlines={circle, magnification=3, connect spies}]
    \node { \includegraphics[width=4cm]{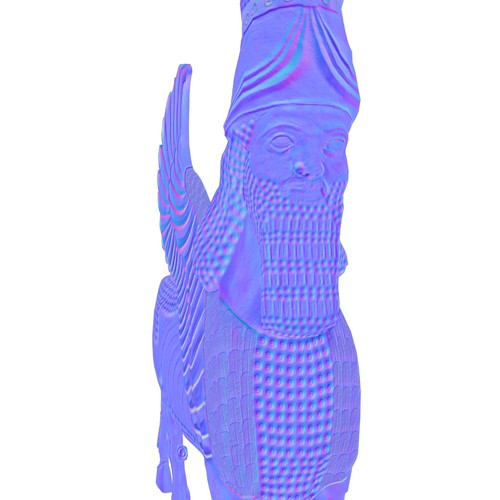} };
    \spy[size=1.25cm] on (-0.5, -1.0) in node at (1.35, -0.5);
    \spy[size=1.25cm] on (-0.6, 0.5) in node at (1.35, 1.5);
\end{tikzpicture}
&
\begin{tikzpicture}[spy using outlines={circle, magnification=3, connect spies}]
    \node { \includegraphics[width=4cm]{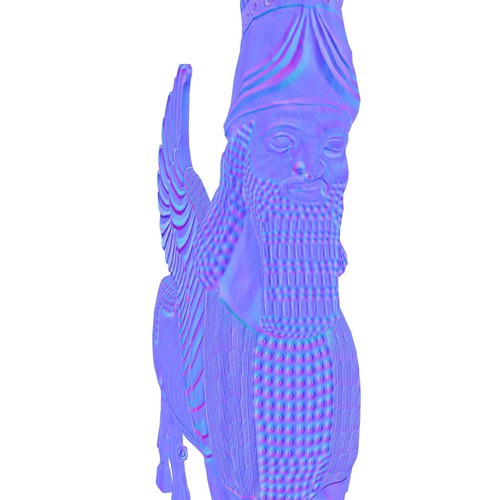} };
    \spy[size=1.25cm] on (-0.5, -1.0) in node at (1.35, -0.5);
    \spy[size=1.25cm] on (-0.6, 0.5) in node at (1.35, 1.5);
\end{tikzpicture}
\end{tabular}
}
\end{center}
\titlecaptionof{figure}{Bump map}{
    Similar surface bumps in world space (left) are dissimilar in the UV tangent space (middle) because of the arbitrary UV mapping.
    Representing the bump map in a tangent space solely dependent on the geometry (right) resolves this issue.
}\label{fig:markspace_normal}

%% file: fig/static_illumination.tex
\begin{center}
\resizebox{\linewidth}{!}{
\renewcommand{\arraystretch}{0.6}%
\setlength{\tabcolsep}{0pt}%
\begin{tabular}{%
    @{}>{\centering\arraybackslash}m{5cm}%
    >{\centering\arraybackslash}m{5cm}%
    @{}
}
Viewpoint 1 & Viewpoint 2 \\
\includegraphics[width=5cm]{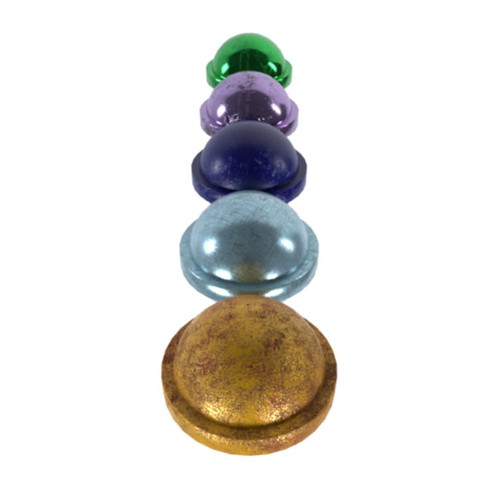} &
\includegraphics[width=5cm]{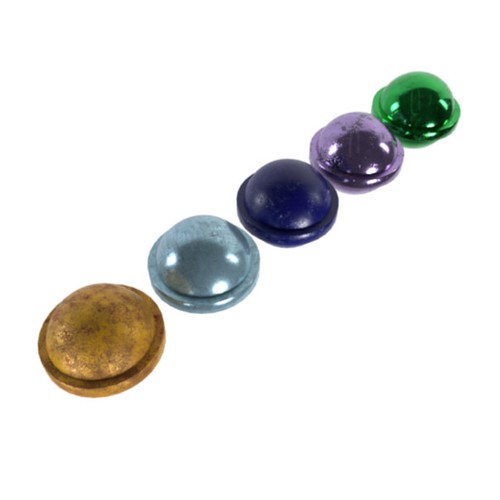}
\end{tabular}
}
\end{center}
\titlecaptionof{figure}{Rendering function}{The dataset is constructed so that the lighting remains constant with respect to the camera, simplifying the rendering function $f_{RGB}$: notice the similar highlight location.
}\label{fig:camera_colocated_env_illumination}

%% file: img/tikz/control_diagram_tikz.tex
\begin{center}
    \begin{minipage}{0.235\textwidth}
        \centering
        \include{img/tikz/control_diagram_parts/control_net}
        (a) ControlNet
    \end{minipage}
    \hfill
    \begin{minipage}{0.235\textwidth}
        \centering
        \include{img/tikz/control_diagram_parts/control_net_xs}
        (b) ControlNet-Xs
    \end{minipage}
    \hfill
    \begin{minipage}{0.235\textwidth}
        \centering
        \include{img/tikz/control_diagram_parts/animate_anyone}
        (c) AnimateAnyone
    \end{minipage}
    \hfill
    \begin{minipage}{0.235\textwidth}
        \centering
        \include{img/tikz/control_diagram_parts/collaborative_control}
        (d) Proposed
    \end{minipage}
    \end{center}
    \caption{
    High-level overview of communication in (a) ControlNet~\cite{zhang2023adding}, (b) ControlNet-XS~\cite{zavadski2023controlnetxs}, (c) AnimateAnyone~\cite{hu2023animateanyone} and (d) our proposed Collaborative Control approach.
    Blue represents frozen blocks, while orange elements are optimized during training.
    }\label{fig:control_diagram}

%% file: sec/4_results.tex
\section{Results}\label{sec:results}
\inlinesection{Distribution match metrics} As an evaluation of how well the data distribution is modeled, distribution match is considered a proxy to both quality and diversity.
The Inception Score (IS~\cite{salimans2016improved}), which checks the distribution match against ImageNet, is not relevant in a PBR context as it applies only to RGB images.
The Fréchet Inception Distance (FID~\cite{heusel2017gans}), which compares the distributions of the last hidden state of the Inceptionv3~\cite{szegedy2016rethinking} network on both real and generated images, has been found to better align to perceptual quality.
Finally, the recently introduced CLIP Maximum-Mean Discrepancy (CMMD~\cite{jayasumana2023rethinking}) compares the distribution of the CLIP embeddings of generated images to that of a reference dataset.
It offers significantly improved sample efficiency, and was shown by the authors to be a better indicator of low-level image quality than FID.
However, as these metrics are intended for three-channel color images, we evaluate them on PBR images following Chambon et al.~\cite{chambon2021passing}, by averaging the relevant scores of multiple triplets.
We report as PBR distribution match the average of the scores over each of the PBR channels independently, as well as over three additional triplets, as the full set of triplets is prohibitively expensive to compute (the supplementary contains all the constituting scores).
The additional triplets are (grayscale albedo, roughness, metallic), (roughness, metallic, normal XY norm) and (grayscale albedo, normal X, normal Y) for a balance between the full cartesian product (which is costly) and mixing channels that are normally relatively independent.

\inlinesection{Out-of-distribution (OOD) performance metrics} indicate the level to which our generator can align to conditioning that it was not trained on. Recent work has introduced the CLIP alignment score~\cite{hessel2021clipscore, foong2023challenges}, which estimates the average distance between the text prompt CLIP embedding and the generated image's CLIP embedding, indicating how faithfully the prompt was followed.
Additionally, OneAlign~\cite{wu2023qalign} is a neural model that estimates aesthetics and quality scores for images, shown to align well with human opinions, summarized in a QAlign score for both aesthetics and general quality.
In order to evaluate the OOD performance, we randomly select 50 objects from Objaverse, and generate unlikely appearance prompts for them using ChatGPT4~\cite{achiam2023gpt}.
These results, as well as a t-SNE plot of the embeddings of the original and OOD prompts, are integrally shown in the supplementary material.

\subsection{Comparisons and Ablations}

To the best of our knowledge, there are no published PBR generation models that generate PBR images for entire objects or scenes (only for generation of single materials~\cite{vecchio2023controlmat,Sartor2023Matfusion}).
Therefore, we perform an extensive ablation study on our design choices, taking care to include typical approaches from techniques that generate other modalities than PBR.
Please refer to \cref{fig:qualitative_results_all} for the qualitative comparisons, while \cref{tab:quantitative_results_compact} contains quantitative results.

\inlinesection{Comparison between control paradigms}
We compare the performance of the proposed bidirectional cross-network communication layer against two other paradigms: one inspired by ControlNet-XS~\cite{zavadski2023controlnetxs}, and one inspired by AnimateAnyone~\cite{hu2023animateanyone}.
In the former, dubbed \emph{one-way} communication, the communication layers receive as input only the RGB model's internal state, and they only affect the PBR model's internal state.
The latter, dubbed \emph{clockwise} communication, functions in the same way for the encoder part of the architecture, but reverses the information flow to go from the PBR model to the RGB model for the decoder half of the architecture.
We see that the \emph{one-way} attention does not perform well, with lower distribution match scores as well as OOD performance scores; the frozen RGB model cannot realign to the conditional distribution required from it in \cref{eq:joint_to_parallel_diffusion}, and we see that the positions it generates for the objects does not align at all with the mask from the normal image.
The \emph{clockwise} attention performs significantly better, but is likely still hampered by $\vect{z}^{\prime}_{rgb, t-1}$ not being easily available to the PBR model --- a similar reasoning as to why the authors of ControlNet-XS included the direct communication link between the base and controlling models' encoders.

\inlinesection{Comparison between communication types}
In terms of the type of communication, we compare the proposed single-layer per-pixel communication against a per-pixel MLP-based communication layer, and a global attention layer.
The latter performs surprisingly well considering that it lacks pixel correspondences; it is hard to enforce pixel-wise alignment through a global attention layer, which we hypothesize to be the reason for the lower quantitative performance.
As Jin et al.~\cite{jin2023training} discuss, an attention-based architecture is also less robust to resolution changes.
The per-pixel MLP, containing four hidden per-pixel linear layers with normalization layers~\cite{ba2016layer} in-between, does not qualitatively perform notably better than the single-layer communication layer, so that we settle for the simpler and more computationally efficient choice.

\inlinesection{Comparison against fine-tuning}
We also compare Collaborative Control against the alternative where we edit the first and last layers of the pre-trained network to match the dimensionality of the PBR images (optionally with the rendered image), and then fine-tune the entire network end-to-end.
Although the distribution match scores for these fine-tuning variants are similar to Collaborative Control, the fine-tuning methods strongly overfit to the training data and perform very poorly in a qualitative OOD comparison.

\inlinesection{PBR-specific VAE vs RGB VAE}
We compare the performance of Collaborative Control with a PBR-specific VAE against a version that uses the triplets-based RGB VAE mentioned in \cref{sec:method} to encode the PBR channels (encoding albedo, roughness$+$metallic, and bump maps in separate triplets and concatenating their latent representations).
The mismatch within the PBR domain is clear, both quantitatively through the worse distribution matching scores, and qualitatively in the produced images.

\begin{figure}
    \input{img/qualitative_figure.tex}\label{fig:qualitative_results_all}
\end{figure}

\begin{table}[t]
    \input{tab/ablations_quantitative.tex}
\end{table}

\inlinesection{Impact of the training budget}
Comparing the version training on a single A100 with the version trained with $8$ A100s (for eight times the batch size), we see that the latter performs significantly better quantitatively in terms of distribution match, but not quality.
Visually, the differences are less clear, although the higher-budget model appears to follow complex prompts slightly better.

\inlinesection{Impact of the training resolution}
We compare the performance of Collaborative Control with two training resolutions: $256\times256$ and $512\times512$, both evaluated on $512\times512$ (ZeroDiffusion's native resolution).
While the low-resolution model quantitatively performs better, visually it is clear that it does not capture the same level of detail as the high-resolution model --- we explain this through the metrics focusing on high-level encoding of the images, and the lower resolution enables smoother training through a larger batch size ($42$).

\inlinesection{Impact of training dataset size}
Now, we evaluate the performance of Collaborative Control under data sparsity by evaluating models trained on $98\%$, $20\%$, $5\%$ and $1\%$ of the full $6$M training images.
The proposed approach proves very data-efficient and performs well even when trained on only a few thousand images ($1\%$).
We observe that it is necessary to disable prompt cross-attention in the PBR model, and that this effect gets more pronounced with fewer data: we hypothesize that the model overfits to the training data and that forcing prompt attention to occur through the frozen RGB base model prevents this overfitting.

\inlinesection{Compatibility with other control techniques}
As a closing experiment, we illustrate that Collaborative Control is compatible with other control techniques~\cite{zavadski2023controlnetxs, mou2023t2i, ye2023ip-adapter}, which drastically expands the practical applications of our proposed method.
We demonstrate this specifically with IP-Adapter~\cite{ye2023ip-adapter}, which allows us to condition the final output on a style image by introducing additional style cross-attention layers within the base model.
We can apply an IP-Adapter to the base model and still generate PBR content, as illustrated in \cref{fig:ipadapter_demo_new}.

\begin{figure}[t]
    \centering
    \begin{minipage}[b]{0.48\linewidth}
        \input{fig/relighting.tex}
    \end{minipage}
    \hfill
    \begin{minipage}[b]{0.48\linewidth}
        \input{fig/interpolation.tex}
    \end{minipage}
\end{figure}

\begin{figure}[t]
    \centering
    \begin{minipage}[b]{0.48\linewidth}
    \input{fig/IPAdapter_demo}
    \end{minipage}
    \hfill
    \begin{minipage}[b]{0.48\linewidth}
        \input{fig/failure_cases}
    \end{minipage}
\end{figure}

\inlinesection{Relighting}
For a small qualitative indication that the PBR materials we generate also look natural under different environment lighting, \cref{fig:relighting_environment_influence} shows a generated texture when relit in two novel environment maps. Furthermore, a slice of the lightfield under rotation of the second environment map shows that the highlights and shadows move smoothly (hinting that the bump map is meaningful).

\inlinesection{Interpolation}
To illustrate the stability of our proposed approach in terms of both initial noise and the text prompt, we (separately) interpolate between two prompts and between two initial noises in \cref{fig:space_stability}.
We perform prompt interpolation in CLIP space, and for noise interpolation we rescale the final image to have unit standard deviation and zero mean after blending.
The resulting images appear natural yet meaningfully interpolate between both extremes.

\subsection{Limitations and failure cases}
We identify two major failure cases: lack of detail in the roughness, metallic, and bump maps, and a failure to follow OOD prompts.
In the former, we see (\eg\cref{fig:failure_cases}) that the model outputs a constant (though varying per instance) roughness and metallic value, and a flat bump map.
We attribute this to the training data: Objaverse contains many objects with constant roughness and metallic properties and a flat bump map --- likely biasing the model towards such outputs.
Anecdotally, we have found that selecting a different random seed will often succeed where the first generation disappointed --- practically, the model produces very diverse results even for the same prompt and the same conditioning geometry, so that we argue that this is either not a significant issue or can be resolved with better training data.

A failure to follow out-of-distribution prompts happens mostly when structural features in the prompt are incompatible with the conditioning geometry, such as for example \emph{a gilded lion} for a table mesh.
We hypothesize that the control signal from the PBR model conflicts with the text cross-attention in the frozen RGB model, resulting in lackluster outputs. Different random seeds occasionally resolve this issue, albeit more rarely.

Finally, we note that the model does come at the cost of executing two parallel diffusion models.
We note that the motivation for this was mainly to retain a frozen copy of the base RGB model: in applications where these requirements are prohibitively expensive, distilling our approach into a direct PBR model is likely to bring relief.

%% file: img/qualitative_figure.tex
\begin{center}
\begin{minipage}[r]{0.12\linewidth}\flushright{}
    \hfill
\end{minipage}
\hfill
\begin{minipage}{0.28\linewidth}\hfill
    \begin{minipage}{0.33\linewidth}
        \includegraphics[width=\linewidth, height=\linewidth]{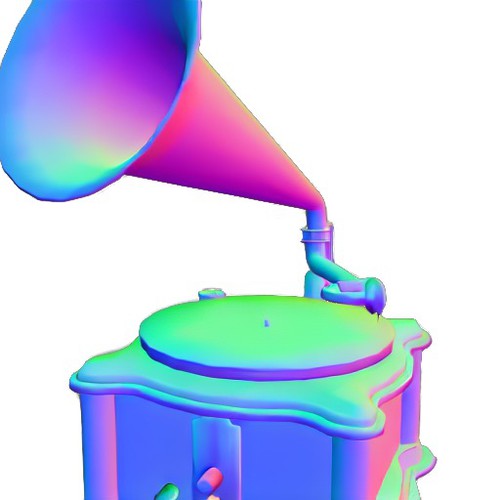}
    \end{minipage}
    \hfill
    \begin{minipage}{0.28\linewidth}\centering\tiny
        In domain
    \end{minipage}
    \hfill
    \begin{minipage}{0.33\linewidth}
        \includegraphics[width=\linewidth, height=\linewidth]{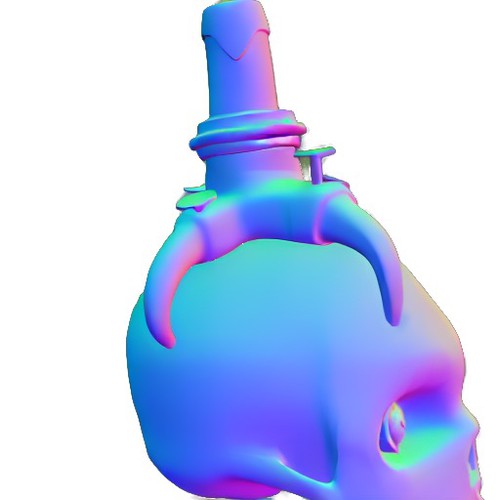}
    \end{minipage}
\end{minipage}
\hfill
\begin{minipage}{0.28\linewidth}\hfill
    \begin{minipage}{0.55\linewidth}\flushright{}\tiny A steam-punk bronze robot dog with a globe on its back and an ornate Victorian head\end{minipage}
    \begin{minipage}{0.33\linewidth}
        \includegraphics[width=\linewidth, height=\linewidth]{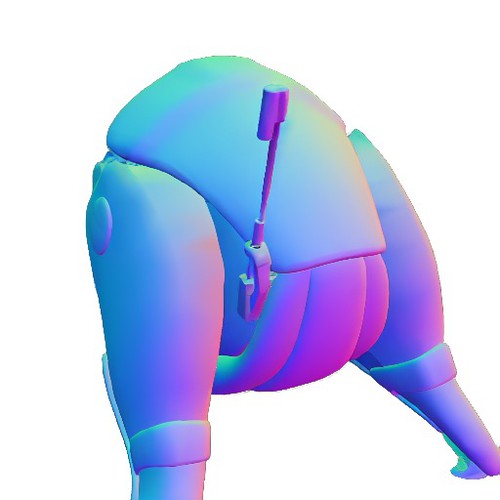}
    \end{minipage}
\end{minipage}
\hfill
\begin{minipage}[c]{0.28\linewidth}\hfill
    \begin{minipage}{0.55\linewidth}\flushright{}\tiny A crystalline vintage movie camera shimmering with prismatic light\end{minipage}
    \begin{minipage}{0.33\linewidth}
        \includegraphics[width=\linewidth, height=\linewidth]{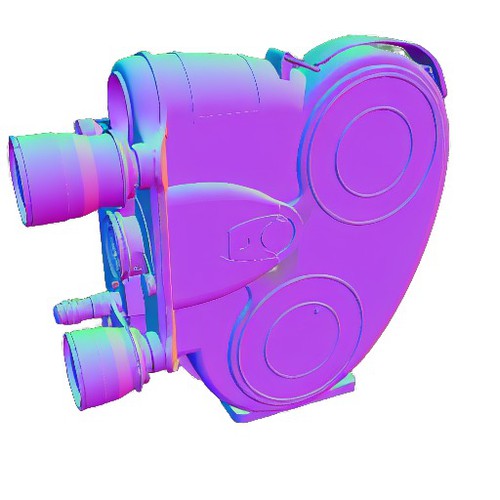}
    \end{minipage}
\end{minipage}

\begin{minipage}[r]{0.12\linewidth}\flushright{}
    \tiny
        Proposed

        (Zero-

        Diffusion

        as base

        model)
\end{minipage}
\hfill
\begin{minipage}[c]{0.28\linewidth}
    \centering
    \includegraphics[width=\linewidth]{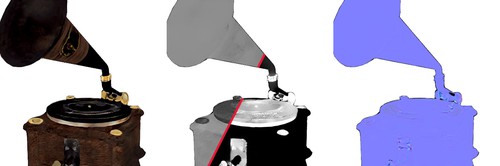}
    \includegraphics[width=\linewidth]{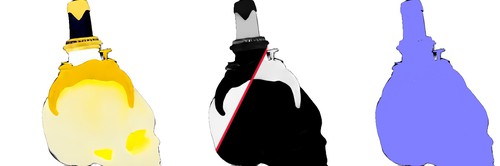}
\end{minipage}
\hfill
\begin{minipage}[c]{0.28\linewidth}
    \centering
    \includegraphics[width=\linewidth]{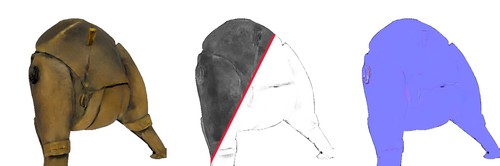}
    \includegraphics[width=\linewidth]{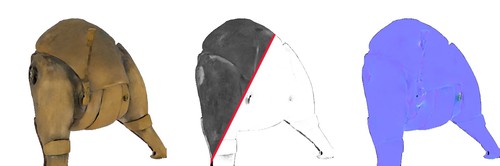}
\end{minipage}
\hfill
\begin{minipage}[c]{0.28\linewidth}
    \centering
    \includegraphics[width=\linewidth]{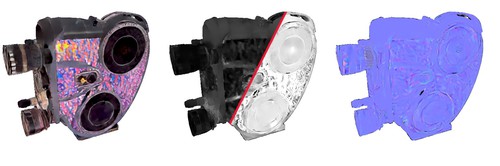}
    \includegraphics[width=\linewidth]{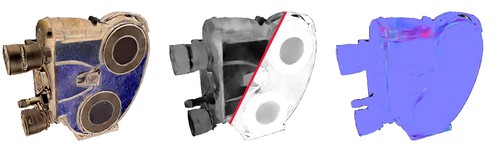}
\end{minipage}

\begin{minipage}[r]{\linewidth}
    \hfill \hbox to 0.90\textwidth{\leaders\hbox to 5pt{\hss - \hss}\hfil}
\end{minipage}

\begin{minipage}[r]{0.12\linewidth}\flushright{}
    \tiny
        Proposed

        (higher

        training

        budget)
\end{minipage}
\hfill
\begin{minipage}[c]{0.28\linewidth}
    \centering
    \includegraphics[width=\linewidth]{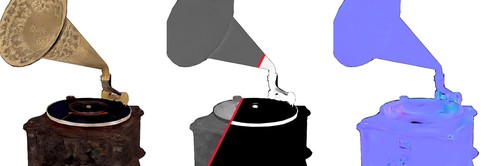}
    \includegraphics[width=\linewidth]{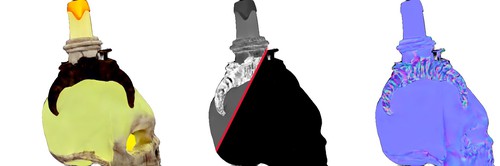}
\end{minipage}
\hfill
\begin{minipage}[c]{0.28\linewidth}
    \centering
    \includegraphics[width=\linewidth]{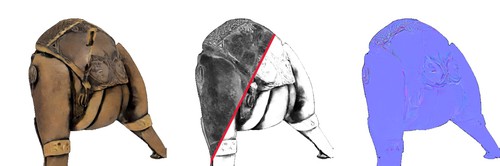}
    \includegraphics[width=\linewidth]{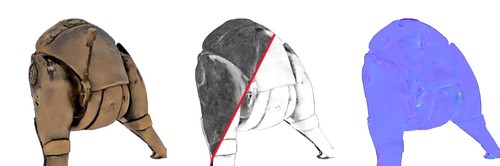}
\end{minipage}
\hfill
\begin{minipage}[c]{0.28\linewidth}
    \centering
    \includegraphics[width=\linewidth]{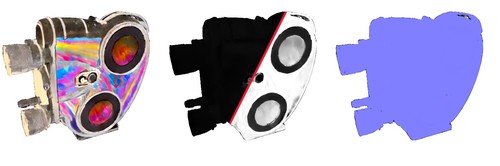}
    \includegraphics[width=\linewidth]{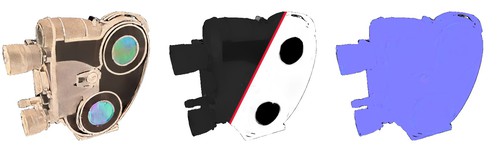}
\end{minipage}

\begin{minipage}[r]{\linewidth}
    \hfill \hbox to 0.90\textwidth{\leaders\hbox to 5pt{\hss - \hss}\hfil}
\end{minipage}

\begin{minipage}[r]{0.12\linewidth}\flushright{}
    \tiny
        Fine-tuning

        the base

        model
\end{minipage}
\hfill
\begin{minipage}[c]{0.28\linewidth}
    \centering
    \includegraphics[width=\linewidth]{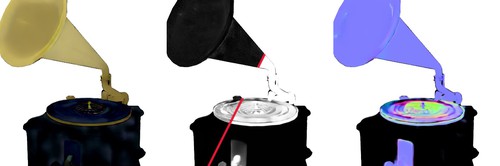}
    \includegraphics[width=\linewidth]{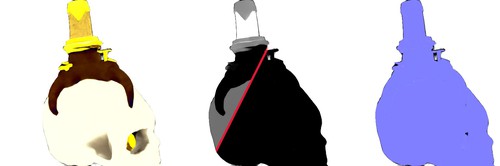}
\end{minipage}
\hfill
\begin{minipage}[c]{0.28\linewidth}
    \centering
    \includegraphics[width=\linewidth]{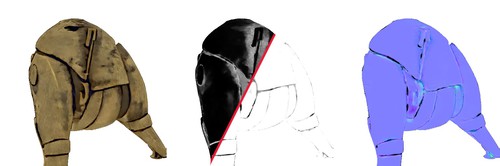}
    \includegraphics[width=\linewidth]{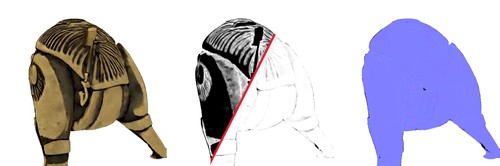}
\end{minipage}
\hfill
\begin{minipage}[c]{0.28\linewidth}
    \centering
    \includegraphics[width=\linewidth]{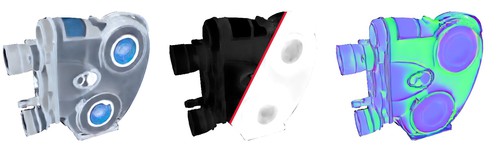}
    \includegraphics[width=\linewidth]{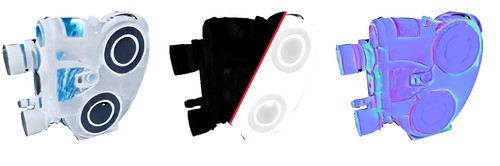}
\end{minipage}

\begin{minipage}[r]{\linewidth}
    \hfill \hbox to 0.90\textwidth{\leaders\hbox to 5pt{\hss - \hss}\hfil}
\end{minipage}

\begin{minipage}[r]{0.12\linewidth}\flushright{}
    \tiny
        Trained on

        $256\times256$
\end{minipage}
\hfill
\begin{minipage}[c]{0.28\linewidth}
    \centering
    \includegraphics[width=\linewidth]{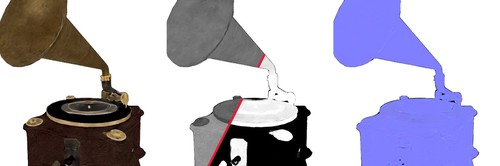}
    \includegraphics[width=\linewidth]{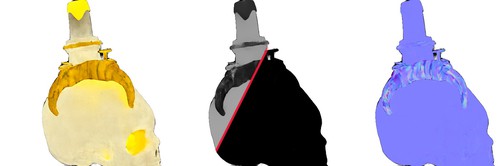}
\end{minipage}
\hfill
\begin{minipage}[c]{0.28\linewidth}
    \centering
    \includegraphics[width=\linewidth]{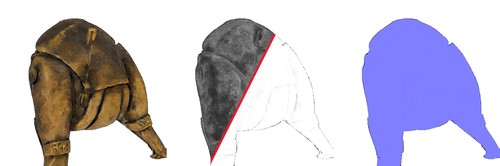}
    \includegraphics[width=\linewidth]{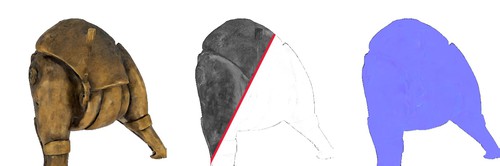}
\end{minipage}
\hfill
\begin{minipage}[c]{0.28\linewidth}
    \centering
    \includegraphics[width=\linewidth]{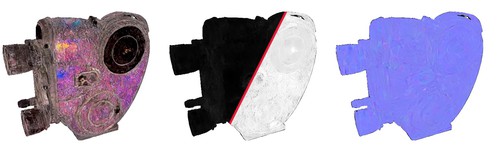}
    \includegraphics[width=\linewidth]{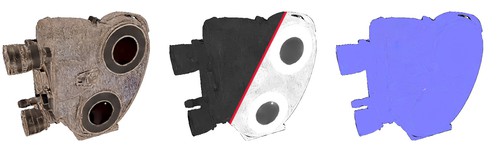}
\end{minipage}

\begin{minipage}[r]{\linewidth}
    \hfill \hbox to 0.90\textwidth{\leaders\hbox to 5pt{\hss - \hss}\hfil}
\end{minipage}

\begin{minipage}[r]{0.12\linewidth}\flushright{}
    \tiny
        1\% of data

        PBR prompt

        attention
\end{minipage}
\hfill
\begin{minipage}[c]{0.28\linewidth}
    \centering
    \includegraphics[width=\linewidth]{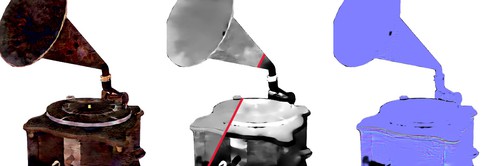}
    \includegraphics[width=\linewidth]{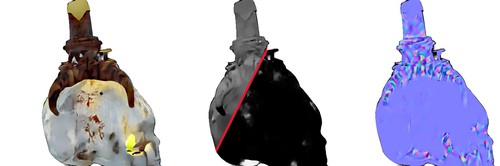}
\end{minipage}
\hfill
\begin{minipage}[c]{0.28\linewidth}
    \centering
    \includegraphics[width=\linewidth]{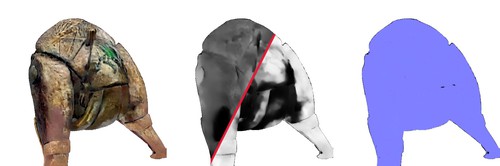}
    \includegraphics[width=\linewidth]{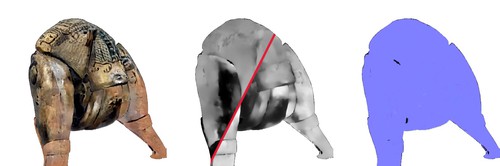}
\end{minipage}
\hfill
\begin{minipage}[c]{0.28\linewidth}
    \centering
    \includegraphics[width=\linewidth]{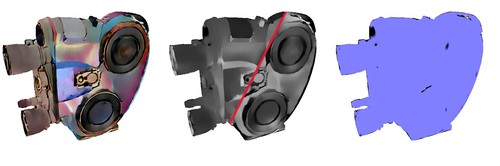}
    \includegraphics[width=\linewidth]{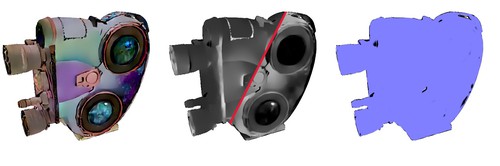}
\end{minipage}

\begin{minipage}[r]{\linewidth}
    \hfill \hbox to 0.90\textwidth{\leaders\hbox to 5pt{\hss - \hss}\hfil}
\end{minipage}

\begin{minipage}[r]{0.12\linewidth}\flushright{}
    \tiny
        1\% of data

        No PBR prompt

        attention
\end{minipage}
\hfill
\begin{minipage}[c]{0.28\linewidth}
    \centering
    \includegraphics[width=\linewidth]{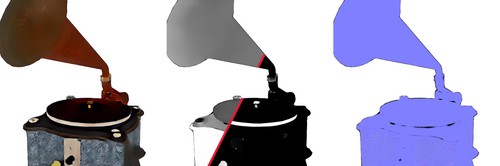}
    \includegraphics[width=\linewidth]{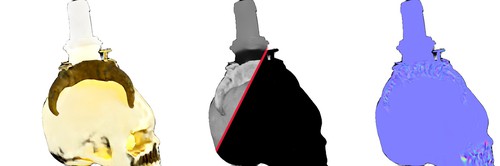}
\end{minipage}
\hfill
\raisebox{0pt}[0pt][0pt]{\vrule height 15.1cm depth 1.1cm width 0.5pt}
\begin{minipage}[c]{0.28\linewidth}
    \centering
    \includegraphics[width=\linewidth]{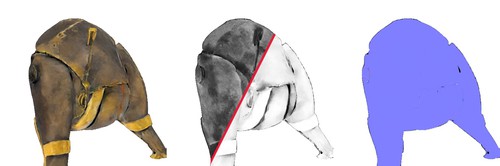}
    \includegraphics[width=\linewidth]{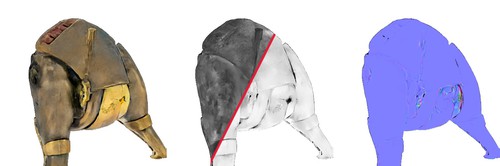}
\end{minipage}
\hfill
\begin{minipage}[c]{0.28\linewidth}
    \centering
    \includegraphics[width=\linewidth]{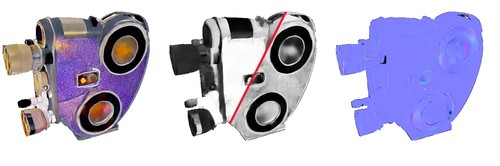}
    \includegraphics[width=\linewidth]{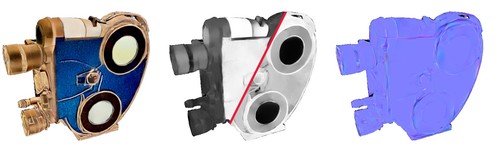}
\end{minipage}

\end{center}
\caption{
    Generated albedo, roughness/metallic and bump map images from the ablation studies.
    While significant quality differences are visible, only the fine-tuning approach and the data-sparse regime \emph{with} PBR prompt cross-attention fail completely.
    The version that was trained on a smaller resolution does not break but does not result in maximum detail either.
    Best viewed digitally.
}

%% file: tab/ablations_quantitative.tex
\caption{%
Quantative results for all evaluated variants. The ablation baseline is highlighted in bold, duplicated for easier comparisons within the individual ablations.%
}\label{tab:quantitative_results_compact}
\begin{adjustbox}{width=\textwidth}
    \begin{tabular}{lcr|cc:cc:cc:cc|cc:cc:cc}
    \multicolumn{1}{l}{}                 &                                                              & \multicolumn{1}{l|}{} & \multicolumn{8}{c|}{2\% held-out evaluation data}                                                                                                                                                                                                                                                                                                   & \multicolumn{6}{c}{OOD}                                                                                                                                                                                                                                            \\
    \multicolumn{1}{l}{}                 &                                                              & \multicolumn{1}{l|}{} & \multicolumn{2}{c:}{\multirow{2}{*}{CMMD $\downarrow$}}                                         & \multicolumn{2}{c:}{\multirow{2}{*}{FID $\downarrow$}}                                          & \multicolumn{4}{c|}{QAlign $\uparrow$}                                                                                                                                                 & \multicolumn{4}{c:}{QAlign $\uparrow$}                                                                                                                                                  & \multicolumn{2}{c}{\multirow{2}{*}{CLIPScore $\uparrow$}}                                       \\
    \multicolumn{1}{l}{}                 &                                                              & \multicolumn{1}{l|}{} & \multicolumn{2}{c:}{}                                                              & \multicolumn{2}{c:}{}                                                              & \multicolumn{2}{c:}{Ae}                            & \multicolumn{2}{c|}{Q}                            & \multicolumn{2}{c:}{Ae}                            & \multicolumn{2}{c:}{Q}                             & \multicolumn{2}{c}{}                                                                 \\
    \multicolumn{1}{l}{}                 &                                                              & \multicolumn{1}{l|}{} & \rotatebox{90}{PBR} & \rotatebox{90}{Relit} & \rotatebox{90}{PBR} & \rotatebox{90}{Relit} & \rotatebox{90}{Albedo} & \rotatebox{90}{Relit} & \rotatebox{90}{Albedo} & \rotatebox{90}{Relit} & \rotatebox{90}{Albedo} & \rotatebox{90}{Relit} & \rotatebox{90}{Albedo} & \rotatebox{90}{Relit} & \rotatebox{90}{Albedo} & \rotatebox{90}{Relit} \\ \hline
    \multirow{6}{*}{\parbox{0.95in}{\vspace{5.5pt}\smash{Communication}}}       & \multicolumn{2}{r|}{one-way}                                                                                             & 16.44 & 13.38 & 20.90 & 16.39 & 1.95 & 1.97 & 2.37 & 2.48 & 1.91 & 1.59 & 2.35 & 1.69                                                                                                         & 23.08 & 23.40                                                      \\
                                         & \multicolumn{2}{r|}{clockwise}                                                                                           & 6.78  & 2.76  & 12.21 & 11.53 & 2.04 & 2.02 & 2.63 & 2.64 & \bestcolor{2.14} & 1.70 & 2.77 & \bestcolor{1.74}                                                                                 & 26.45 & 24.53                                    \\
                                         & \multicolumn{2}{r|}{\textbf{bi-directional}}                                                                             & \bestcolor{6.30}  & \bestcolor{1.79}  & \bestcolor{11.65} & \bestcolor{10.64} & \bestcolor{2.11} & 2.03 & \bestcolor{2.75} & \bestcolor{2.66} & 2.12 & 1.76 & \bestcolor{2.78} & 1.73         & \bestcolor{26.76} & \bestcolor{25.41}                                    \\ \cdashline{2-17} 

                                         & \multicolumn{2}{r|}{\textbf{Pixel-wise zero-conv}}                                                                       & 6.30  & \bestcolor{1.79}  & 11.65 & \bestcolor{10.64} & \bestcolor{2.11} & \bestcolor{2.03} & \bestcolor{2.75} & \bestcolor{2.66} & 2.12 & 1.76 & 2.78 & 1.73                                             & 26.76 & 25.41                                    \\
                                         & \multicolumn{2}{r|}{Pixel-wise MLP}                                                                                      & \bestcolor{5.43}  & 1.87  & \bestcolor{11.43} & 10.67 & 2.10 & 2.02 & 2.74 & \bestcolor{2.66} & \bestcolor{2.26} & 1.75 & \bestcolor{2.96} & \bestcolor{1.81}                                 & \bestcolor{27.15} & \bestcolor{25.95}                                    \\
                                         & \multicolumn{2}{r|}{Global Attention}                                                                                    & 7.60  & 5.22  & 13.61 & 11.93 & 1.94 & 1.98 & 2.51 & 2.60 & 1.99 & 1.72 & 2.71 & 1.80                                                                                                         & 24.50 & 24.01                                    \\ \hline

    \multicolumn{3}{r|}{\textbf{Collaborative Control}}                                                                                                             & 6.30  & \bestcolor{1.79}  & 11.65 & \bestcolor{10.64} & \bestcolor{2.11} & 2.03 & \bestcolor{2.75} & \bestcolor{2.66} & 2.12 & 1.76 & 2.78 & 1.73                                             & \bestcolor{26.76} & \bestcolor{25.41}                                    \\
    \multicolumn{3}{r|}{Fine-tuning (with RGB output)}                                                                                                               & 13.40 & 2.78  & 14.42 & 10.79 & 2.05 & 2.02 & 2.60 & 2.61 & 2.10 & 1.76 & 2.62 & \bestcolor{1.86}                                                                                             & 25.04 & 22.69                                    \\
    \multicolumn{3}{r|}{Fine-tuning (without RGB output)}                                                                                                            & \bestcolor{5.25}  & 2.88  & \bestcolor{11.41} & 11.37 & 2.03 & 1.99 & 2.58 & 2.58 & \bestcolor{2.26} & 1.71 & \bestcolor{2.97} & 1.81                                                         & 25.66 & 23.31                                    \\ \hline

    \multicolumn{3}{r|}{\textbf{PBR VAE}}                                                                                                                           & \bestcolor{6.30}  & \bestcolor{1.79}  & \bestcolor{11.65} & \bestcolor{10.64} & 2.11 & \bestcolor{2.03} & \bestcolor{2.75} & \bestcolor{2.66} & 2.12 & 1.76 & 2.78 & 1.73                     & \bestcolor{26.76} & \bestcolor{25.41}                                    \\
    \multicolumn{3}{r|}{RGB VAE on triplets}                                                                                                                        & 84.66 & 5.99  & 25.81 & 11.63 & \bestcolor{2.16} & 1.99 & 2.67 & 2.55 & \bestcolor{2.30} & 1.71 & \bestcolor{2.95} & \bestcolor{1.80}                                                         & 25.27 & 23.98                                     \\ \hline

    \multirow{3}{*}{\parbox{0.5in}{\smash{Training budget}}}     & \multicolumn{2}{r|}{\textbf{1 A100, two days}}                                                                           & 6.30  & 1.79  & 11.65 & 10.64 & \bestcolor{2.11} & 2.03 & \bestcolor{2.75} & 2.66 & \bestcolor{2.12} & 1.76 & 2.78 & 1.73                                                                     & 26.76 & \bestcolor{25.41}                                                        \\
                                         & \multicolumn{2}{r|}{8 A100s, two days}                                                                                   & \bestcolor{2.96}  & \bestcolor{1.12}  & \bestcolor{9.55}  & \bestcolor{9.76}  & 2.08 & 2.04 & 2.68 & 2.67 & 2.01 & 1.73 & \bestcolor{2.82} & \bestcolor{1.82}                                 & \bestcolor{26.78} & 25.22                                                        \\ \hline

    \multirow{2}{*}{\parbox{0.5in}{\smash{Training resolution}}} & \multicolumn{2}{r|}{256$\times$256}                                                                                      & \bestcolor{2.23}  & \bestcolor{1.44}  & \bestcolor{9.82}  & \bestcolor{10.20} & 2.10 & 2.04 & 2.73 & 2.68 & 2.20 & 1.78 & \bestcolor{3.13} & \bestcolor{1.80}                                 & 26.71 & 25.21                                    \\
                                         & \multicolumn{2}{r|}{\textbf{512$\times$512}}                                                                             & 6.30  & 1.79  & 11.65 & 10.64 & \bestcolor{2.11} & 2.03 & \bestcolor{2.75} & 2.66 & \bestcolor{2.12} & 1.76 & 2.78 & 1.73                                                                     & \bestcolor{26.76} & \bestcolor{25.41}                                    \\ \hline
 
    \multirow{8}{*}{\parbox{0.5in}{Training\\data}}       & \multicolumn{1}{c}{\multirow{4}{*}{\shortstack{\textbf{No PBR}\\ \textbf{prompt}\\ \textbf{attention}}}} & 1\%                      & 6.25  & \bestcolor{1.43}  & 11.87 & 10.79 & 2.18 & 2.04 & \bestcolor{2.86} & 2.69 & \bestcolor{2.35} & 1.76 & \bestcolor{3.35} & 1.89                                                         & 26.58 & 25.11                                    \\
                                         & \multicolumn{1}{c}{}                                         & 5\%                                                       & \bestcolor{5.77}  & 1.45  & \bestcolor{11.49} & \bestcolor{10.54} & 2.13 & 2.04 & 2.78 & 2.69 & 2.09 & 1.73 & 3.01 & 1.84                                                                                 & \bestcolor{27.28} & 25.04                                    \\
                                         & \multicolumn{1}{c}{}                                         & 20\%                                                      & 5.97  & 1.68  & 11.50 & 10.61 & 2.12 & 2.03 & 2.78 & 2.67 & 2.23 & 1.76 & 3.23 & 1.88                                                                                             & 25.72 & 24.99                                    \\
                                         & \multicolumn{1}{c}{}                                         & \textbf{98\%}                                             & 6.30  & 1.79  & 11.65 & 10.64 & 2.11 & 2.03 & 2.75 & 2.66 & 2.12 & 1.76 & 2.78 & 1.73                                                                                                         & 26.76 & 25.41                                    \\ \cdashline{2-17}
                                         & \multicolumn{1}{c}{\multirow{4}{*}{\shortstack{PBR prompt\\ attention}}}    & 1\%                                        & 20.61 & 4.25  & 18.35 & 12.16 & \bestcolor{2.19} & 2.03 & 2.76 & 2.62 & 2.25 & 1.61 & 2.83 & 1.80                                                                                             & 24.75 & 23.25                                    \\
                                         & \multicolumn{1}{c}{}                                         & 5\%                                                       & 12.17 & 2.58  & 14.95 & 10.97 & 2.13 & 2.04 & 2.71 & 2.65 & 2.27 & 1.80 & 2.97 & \bestcolor{1.98}                                                                                             & 26.89 & \bestcolor{25.53}                                    \\
                                         & \multicolumn{1}{c}{}                                         & 20\%                                                      & 11.35 & 2.33  & 14.78 & 10.78 & 2.13 & 2.03 & 2.74 & 2.65 & 2.29 & 1.76 & 3.07 & 1.77                                                                                                         & 26.79 & 25.39                                    \\
                                         & \multicolumn{1}{c}{}                                         & 98\%                                                      & 9.18  & 2.57  & 13.25 & 11.02 & 2.10 & 2.03 & 2.68 & 2.64 & 2.17 & 1.80 & 2.84 & 1.87                                                                                                         & 25.80 & 24.83                                   
    \end{tabular}
\end{adjustbox}

%% file: fig/relighting.tex
\centering
\begin{minipage}[b]{0.32\linewidth}
    \includegraphics[width=\linewidth, height=\linewidth]{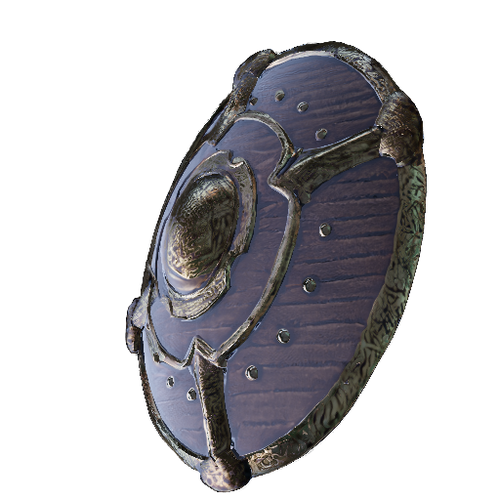}
\end{minipage}
\begin{minipage}[b]{0.32\linewidth}
    \includegraphics[width=\linewidth, height=\linewidth]{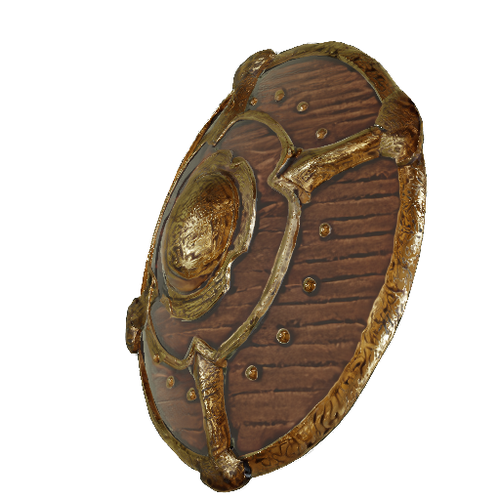} 
\end{minipage}
\hfill
\begin{minipage}[b]{0.32\linewidth}
    \includegraphics[width=0.9\linewidth, height=0.9\linewidth,angle=180,origin=c]{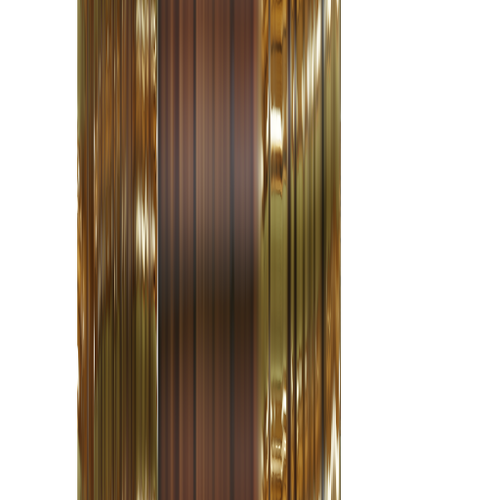} 
\end{minipage}
\caption[]{A re-lit generated texture in \href{https://polyhaven.com/a/peppermint_powerplant_2}{Peppermint Powerplant} and \href{https://polyhaven.com/a/pine_attic}{Pine Attic} and lightfield slice under environment rotation.}\label{fig:relighting_environment_influence}

%% file: fig/interpolation.tex
\centering
\begin{minipage}[c]{0.18\linewidth}
    \centering
    \tiny Van Gogh\\$\rightarrow$ red
    \includegraphics[width=\linewidth, height=\linewidth]{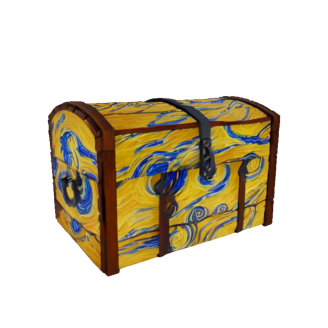} 
    seed 42\\$\rightarrow$ 43
\end{minipage}
\begin{minipage}[c]{0.18\linewidth}
    \centering
    \includegraphics[width=\linewidth, height=\linewidth]{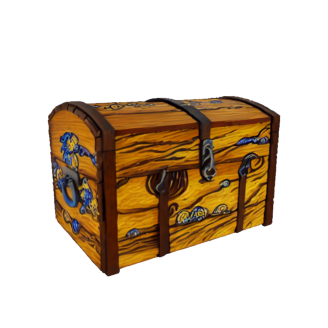}
    \includegraphics[width=\linewidth, height=\linewidth]{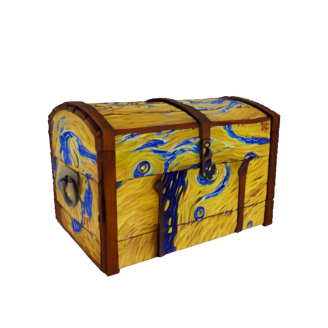}
\end{minipage}
\begin{minipage}[c]{0.18\linewidth}
    \centering
    \includegraphics[width=\linewidth, height=\linewidth]{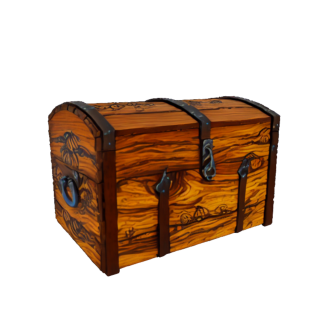}
    \includegraphics[width=\linewidth, height=\linewidth]{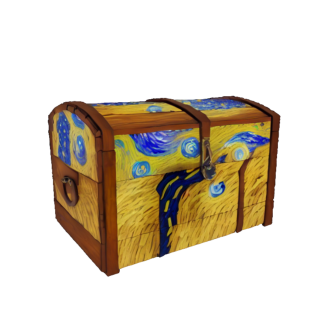}
\end{minipage}
\begin{minipage}[c]{0.18\linewidth}
    \centering
    \includegraphics[width=\linewidth, height=\linewidth]{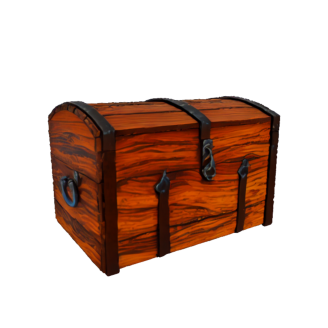}
    \includegraphics[width=\linewidth, height=\linewidth]{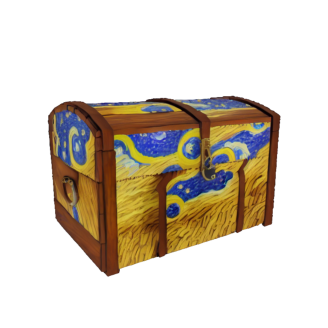}
\end{minipage}
\begin{minipage}[c]{0.18\linewidth}
    \centering
    \includegraphics[width=\linewidth, height=\linewidth]{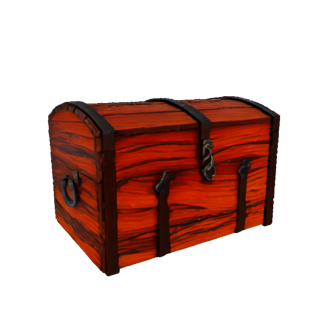}
    \includegraphics[width=\linewidth, height=\linewidth]{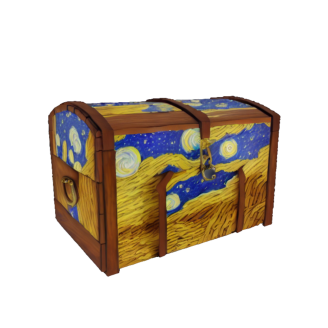}
\end{minipage}
\caption{Interpolation on text embeddings and initial noise shows the stability of the proposed approach in both of these spaces.}\label{fig:space_stability}

%% file: fig/IPAdapter_demo.tex
\begin{center}
\hfill
\begin{minipage}{0.19\linewidth}
    \centering
    \includegraphics[height=\linewidth]{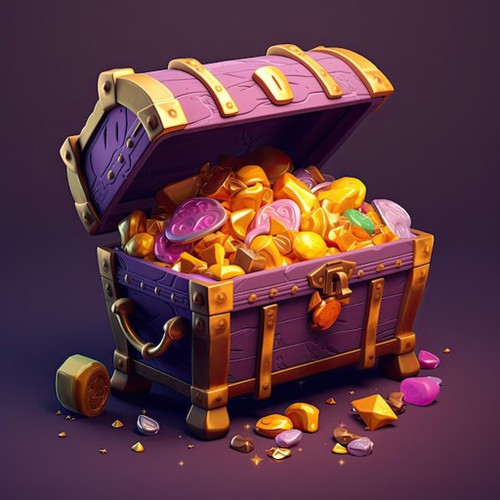}
    \includegraphics[height=\linewidth]{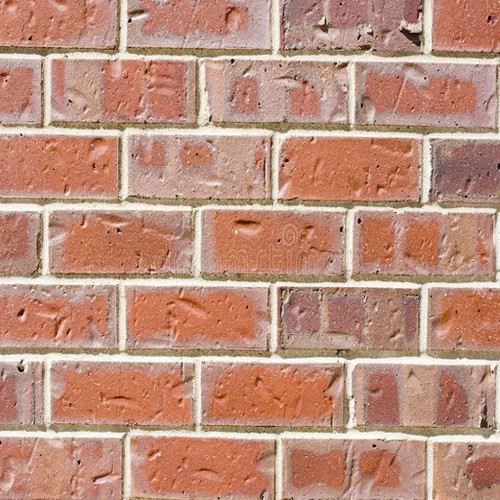}
    \tiny Image Condition
\end{minipage}
\hfill
\begin{minipage}{0.76\linewidth}
    \centering
    \includegraphics[width=\linewidth]{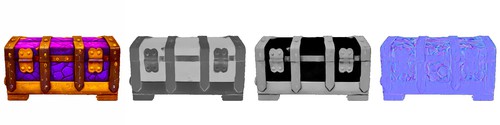}
    \includegraphics[width=\linewidth]{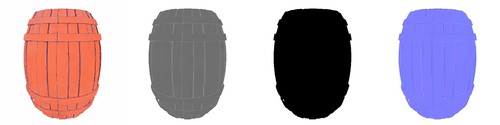}
    \tiny Albedo \hfill Roughness \hfill Metallic \hfill  Bump Map
    
    \phantom{.}
\end{minipage}
\hfill
\end{center}
\caption{
Our PBR diffusion remains compatible with control techniques trained for the base frozen RGB model.
We illustrate this using StableDiffusion 1.5 as the base model using a public IP-Adapter~\cite{ye2023ip-adapter}.}\label{fig:ipadapter_demo_new}

%% file: fig/failure_cases.tex
\begin{center}
\begin{minipage}{0.93\linewidth}
\includegraphics[width=\linewidth]{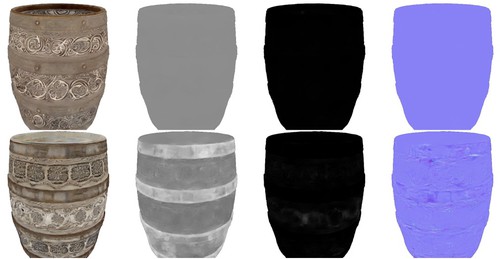}
\end{minipage}
\end{center}
\caption{
Our most common failure case is the constant roughness, metallic or bump maps.
Prompting a \emph{porcelain barrel with intricate designs} for two different random seeds illustrates this behaviour.}\label{fig:failure_cases}

%% file: sec/5_conclusion.tex
\section{Conclusion}
\label{sec:conclusion}
In this work, we have proposed Collaborative Control, a new paradigm for leveraging a pre-trained image-based RGB diffusion model for generating high-quality PBR image content conditioned on object geometry.
We have shown that this bi-directional control paradigm is extremely data-efficient while retaining the high quality and expressiveness of the base RGB model, even when faced with text queries completely out of distribution for the PBR training data.
The plug-and-play nature of our proposed approach is compatible with existing adaptations of the base RGB model, which we have illustrated with IP-Adapter for style guidance of the PBR content.
The availability of high-quality PBR content generation as offered by our proposed approach opens up new avenues for graphics applications, specifically in Text-to-Texture.